\documentclass{article}

\usepackage{arxiv}

\usepackage[utf8]{inputenc} 
\usepackage[T1]{fontenc}    
\usepackage{hyperref}       
\usepackage{url}            
\usepackage{booktabs}       
\usepackage{amsfonts}       
\usepackage{nicefrac}       
\usepackage{microtype}      
\usepackage{lipsum}
\usepackage{xcolor}
\usepackage{blindtext}

\usepackage{todonotes}
\usepackage{booktabs}
\usepackage{multirow}
\usepackage{todonotes}

\usepackage{amsmath,amsfonts,bm}

\def\mF{{\bm{F}}}

\DeclareMathAlphabet{\mathsfit}{\encodingdefault}{\sfdefault}{m}{sl}
\SetMathAlphabet{\mathsfit}{bold}{\encodingdefault}{\sfdefault}{bx}{n}

\def\sZ{{\mathbb{Z}}}

\title{Efficient pre-training objectives for Transformers}

\author{
 Luca Di Liello \\
  DISI Department \\
  University of Trento \\
  Trento 38121, Italy \\
  \texttt{luca.diliello@unitn.it} \\

    \And

  Matteo Gabburo \\
  DISI Department \\
  University of Trento \\
  Trento 38121, Italy \\
  \texttt{matteo.gabburo@unitn.it} \\
   
     \And
 Alessandro Moschitti \\
  Amazon Alexa \\
  Manhattan Beach, CA, USA \\
  \texttt{amosch@amazon.com} \\

}

\begin{document}
\maketitle

\begin{abstract}
Transformer-based neural networks have heavily impacted the field of natural language processing, outperforming most previous state-of-the-art models. However, well-known models such as BERT, RoBERTa, and GPT-2 require a huge compute budget to create a high quality contextualised representations. In this paper, we study several efficient pre-training objectives for Transformers-based models. By testing these objectives on different tasks, we determine which of the ELECTRA model's new features is the most relevant: (i) Transformers pre-training can be improved when the input is not altered with artificial symbols, e.g., masked tokens; and (ii) loss functions computed using the whole output reduce training time. (iii) Additionally, we study efficient models composed of two blocks: a discriminator and a simple generator (inspired by the ELECTRA architecture). Our generator is based on a much simpler statistical approach, which minimally increases the computational cost. Our experiments show that it is possible to efficiently train BERT-like models using a discriminative approach as in ELECTRA but without a complex generator. Finally, we show that ELECTRA largely benefits from a deep hyper-parameter search.

\end{abstract}

\section{Introduction}
\label{sec:intro}
Based on the architecture presented in \cite{vaswani2017attention}, transformer-based models have quickly become the standard in NLP. However, these models require expensive hardware to be pre-trained \cite{strubell-etal-2019-energy}. Recently, there has been many attempts to reduce the computational costs \cite{lan2020albert,sanh2020distilbert,turc2019wellread, clark2020electra}. In particular, we are interested in ELECTRA. This model overcomes the most important BERT's problem of being frequently fed with the MASK token. This causes what is called a pre-training/fine-tuning discrepancy because the MASK token is not seen in fine-tuning. They solve this problem by replacing the Masked Language Model (MLM) objective used by BERT \cite{devlin2019bert} with a simpler objective, which allows the network to be trained as a discriminator instead of a generator. So, instead of predicting original masked tokens, in ELECTRA, the discriminator classifies if tokens are original or fake. Moreover, ELECTRA computes the loss over the entire discriminator output: this can reduce training time because of more content is used from each training example.

We propose several efficient pre-training strategies, and we compare them on well known NLP benchmarks. Specifically, we provide improved versions of MLM and token detection (TD) tasks. For example, we require to predict the token that was replaced instead of simply classifying original and substituted tokens in a binary way. Furthermore, we make the token selection of ELECTRA more efficient by substituting the generator with a simpler generative approaches based on either random selection or history-based statistical model. The latter simply counts the number of time each token has been misclassified by the model. We cluster tokens together to reduce token/count sparsity.

We pre-trained several language models along with state-of-the-art baselines with the same hyper-parameters and datasets. Our models are obtained by combining the above techniques using RoBERTa\cite{liu2019roberta} pre-trained transformer, and we test them on four natural language inference benchmarks: GLUE, SQUAD, ASNQ-R and WikiQA. Moreover, we test the performance on transfer learning, by transferring models on ASNQ-R and by testing them on WikiQA.

Our approaches exhibit a general superiority over MLM. For example, our RoBERTa model that uses a history-based selection of replacement outperforms a similar model based on MLM in most tasks, especially when doing transfer learning, requiring also smaller computational budget. Moreover, our model that only receives replaced tokens and predicts their original value outperforms RoBERTa-MLM in every task by a margin of $1\%$, sometimes matching the performance of ELECTRA, which is a more expensive and refined approach.

Most importantly, (i) in terms of computation time, our generators are much more efficient than the original ones, and (ii) our pre-training objectives require smaller classification heads, thus using less memory and computation resources. We release the source code to reproduce our experiments. \footnote{\url{https://github.com/iKernels/efficient-pre-training-objectives-for-transformers}.}.

\section{Related Work}
\label{sec:rel_work}
Transformer-based models represent a new milestone for the AI community and, in particular, for NLP applications. These models proved to outperform the state of the art in many Natural Language Understanding tasks. The original Transformer architecture was presented in \cite{vaswani2017attention}, where the authors proposed a model based solely on the attention mechanism \cite{luong2015effective}, without using recurrence. This approach had the major benefit of considering very longer sequences at a time, as opposed to RNN. Moreover, Transformers-based models are well suited for transfer learning. In fact, many language models are pre-trained on huge amounts of unlabelled data with different techniques and finally are fine-tuned on the target task. Pre-training is the phase that requires most computational effort, but then the languages models can be shared and fine-tuned on many different tasks in only a few training steps. 

In this paper, we focus on more efficient pre-training objectives. The latter are techniques that enable the training of language models  using unlabelled text. An example is represented by the Causal Language Modeling, exploited by \cite{Radford2018ImprovingLU, radford2019language, brown2020language} for the GPT class of models. CLM consists of providing a model with a truncated sentence and asking the model to predict the next token.

Another significant language modelling objective was proposed in BERT (Bidirectional Encoder Representation for Transformers) \cite{devlin2019bert}. The Masked Language Modeling (MLM) objective consists of masking some tokens of the input sentence and ask the model to predict their original value. By combining the MLM objective with the Next Sentence Prediction loss, the authors of BERT created a state-of-the-art Transformers-based model which, differently from GPT, considers the input in a bidirectional way and is particularly effective for NLU tasks.

Moreover, a relevant language modelling objective was introduced in ELECTRA \cite{clark2020electra}. ELECTRA is composed of a generator and a discriminator network, which are both BERT models. The generator is trained with MLM to find suitable candidates to replace a special MASK token. The discriminator recognizes the tokens replaced by the generator in the original text. After pre-training, the generator is removed, and the discriminator model is used as the pre-trained language model. ELECTRA is able to (i) outperform BERT when trained with the same compute budget or (ii) reach the same performance in much less training time. ELECTRA introduces many innovations, but the most effective is the use of the whole output of the discriminator for computing the loss function. This allows the model to receive a better signal. Additionally, the discriminator input is not affected by the presence of a spurious token such as the MASK token. Indeed, the latter, creating a discrepancy between pre-training and fine-tuning, is one of the main drawbacks of the original BERT.

In the reminder of this section, we present other pre-training techniques that can be used to create lighter and faster models.

In \cite{sanh2020distilbert}, the authors exploit distillation to create a smaller BERT model. Distillation consists of a transfer learning technique, where the knowledge of a large already-trained model is transferred to a smaller model. DistilBERT can achieve $97\%$ of BERT performance by using only about half the computation. A similar technique was exploited in \cite{turc2019wellread}. In this work, the authors train many small models by distillation, studying which of BERT hyper-parameters (e.g., the number of layers, hidden size, embedding size) is the most untactful on the final performance. Even thought the resulting models are very fast in inference because of the reduced number of internal parameters, they are not trained efficiently. Distillation requires a very larger model from which transferring the knowledge to the smaller one.

Finally, ALBERT proposes another relevant methodology that aims at improving pre-training efficiency. The authors of ALBERT \cite{lan2020albert} propose to share the weights of every Transformer layer to save GPU memory and thus be able to use bigger batch sizes. The train a model able to reach state-of-the-art performance in some NLU tasks. Moreover, they show that the contextualized representations created by every Transformer layer are not that different. A very detailed overview of all the recent techniques proposed to improve the efficiency and effectiveness of the Transformer is given in \cite{tay2020efficient}.

In conclusion, many works are addressing the problem of efficient training models. Some apply enhancement at the architectural level, while others design more refined pre-training tasks to train models faster.

\section{Methodology}
\label{sec:method}

This section presents alternative pre-training objectives that can be applied to a wide range of Transformer-based models. Our focus is on efficiency, thus we propose new techniques and/or lighter language models, which can replace inefficient solutions such as, the use of ELECTRA generator, or MLM.

\subsection{Random Token Substitution (RTS)}
\label{ssec:RTS}
Our \textit{Random Token Substitution} approach carries out pre-training technique by learning to discriminate the original tokens from substituted  tokens, similarly to the ELECTRA discriminator. The main difference is that RTS chooses the tokens to be replaced randomly, thus avoiding the use of computational resources to train a separate and expensive generator network. Specifically, the model selects and replaces $15\%$ of the input tokens with others from the vocabulary. Then the modified sentence is provided to the model, which classifies whether tokens are original or not. We apply this technique to a RoBERTa model, and we call it \textbf{RoBERTa-RTS}.

\subsubsection{RTS using aggregated probabilities of misclassifying tokens (C-RTS)}
\label{sec:frequency-generator}

We replace a token $\alpha$ with a token $\beta$ using the posterior misclassification probability, $P(\beta | \alpha \rightarrow \beta)$, where $P$ is the probability that the discriminator is incorrect when classifying $\beta$, given that $\alpha$ was replaced by $\beta$. $P$ can be derived by counting the number of failures/successes for each pair ($\alpha, \beta$) in the previous iterations. Thus, the main difference with the ELECTRA generator is in the context: ELECTRA exploits the whole input sentence to create challenging replacements, while our algorithm only uses the prediction history of single tokens, and only depends on the past predictions of the model over it.

Modelling exactly $P$ is unfeasible given that RoBERTa's vocabulary size of about $50K$ tokens would generate a huge amount of $(\alpha, \beta)$ entries, being also extremely sparse.

Thus, we partition tokens into $n$ different clusters by measuring the Euclidean distance between the corresponding embedding vectors obtained by training a word2vec model \cite{mikolov2013w2v} on the same data used for pre-training. After that, we use the \textit{K-Means} \cite{Lloyd82leastsquares} algorithm to group the tokens into the clusters $C_1, \dots, C_n$.

Our implementation only uses a matrix $\mF \in \sZ^{n \times n}$, which counts the difference between failures and successes of the discriminator for each cluster pairs. $\mF$ is initialised with zeros, then, while training, some tokens are changed into others and the model should discover them. For each pair of tokens $(\alpha \in C_i, \beta \in C_j)$, such that $\alpha$ is replaced with $\beta$, we decrease $\mF_{i,j}$ by $1$ if the model correctly predicts $\alpha \ne \beta$, otherwise we increment it.

For each training step, $P$ is estimated as follows:

\begin{equation}
    P(\beta| \alpha \rightarrow \beta) = P(C_j| C_i \rightarrow C_j) \times P(\beta | C_j)
\end{equation}

assuming that the sampling on the target cluster $C_j$ is done with uniform probability, $P(\beta | C_j) = \frac{1}{|C_j|}$. Then the previous equation becomes:
\begin{equation}
    P(\beta| \alpha \rightarrow \beta) = P(C_j| C_i \rightarrow C_j) \times \frac{1}{|C_j|}
\end{equation}

Thus, we only need to compute $P(C_j | C_i)$ from the counters matrix $\mF$. For each token $\alpha \in C_i$, we define a multinomial distribution over the target clusters by indexing the $\mF_i$ row. To obtain a vector of values that can be interpreted as probabilities, we first apply the min-max normalisation:
$$
\overline{\mF_i} = \frac{\mF_i - \min(\mF_i) }{ \max(\mF_i) - \min(\mF_i) }
$$

and then a $\gamma$-softmax:
$$
P(C_j | C_i) = \frac{ e^{\gamma*\overline{\mF_{i, j}}} }{ \sum_k e^{\gamma*\overline{\mF_{i, k}}} } \\
$$

The $\gamma$ coefficient is used to control how the probability mass is concentrated or relaxed on the most probable cluster. After sampling the target cluster from this multinomial distribution, the model selects the target token randomly with uniform probability.

To summarize, we randomly select some tokens of the input sentence and, given their clusters, we define a multinomial distribution over the target clusters. We finally select tokens with uniform probability from the target clusters and put them in the original sentence. We looked for the best number of clusters $n$ among $\{30, 100, 300, 1000\}$ and for the best $\gamma$ in $\{1, 2, 5, 10\}$. After preliminary experiments we found that the best combination is $n = 100, \gamma = 2$. We named the RoBERTa model exploiting this pre-training technique as \textbf{RoBERTa-C-RTS}.

\subsection{Swapped Language Model (SLM)}
\label{ssec:SLM}
\textit{Swapped Language Model} is a pre-training technique similar to the Masked Language Modeling introduced by BERT\cite{devlin2019bert}. In this case, tokens are only randomly replaced with others and never with the special MASK token. Then, differently from \textit{RTS}, the model is trained to predict the original token and not to discriminate between fakes and originals. This model is still generative but without the need to exploit the MASK token. We apply this technique to RoBERTa and we call it \textbf{RoBERTa-SLM}.

\subsubsection{SLM-all}
\label{ssec:SLM_all}
The \textit{SLM} loss cannot be directly applied to the ELECTRA model. The reason is that \textit{SLM} (like MLM) does not discriminate between original and replaced tokens since it is only applied to the output position corresponding to tampered tokens. Then, we propose an alternative objective called \textit{SLM-all}. In this case, the discriminator has to predict the whole input sentence, estimating which tokens were changed and predicting their original values. At the same time, it should only reproduce the input in output for unchanged inputs. We call this model \textbf{ELECTRA-SLM-all}.

\section{Experiments}
\label{sec:experiments}
This Section describes the datasets, the hyper-parameters and the metrics used both in pre-training and fine-tuning.

\subsection{Models}
To make a fair comparison, we always use the same architecture (corresponding to a RoBERTa-\textit{base} model): $12$ Transformer layers, a hidden size of $768$, $12$ attention heads and an intermediate size equal to $3072$. For ELECTRA, we used a generator with reduced width as described in the original work. Then, it features an intermediate size of $1024$, a hidden-size equal to $256$ and $4$ attention heads and $12$ layers. The discriminator is instead equal to a RoBERTa-\textit{base}. The RoBERTa models contain about 125M parameters, while on the other hand ELECTRA, which is composed of both a generator and a discriminator, has 142M parameters. Please notice that these numbers do not consider the various classification heads used in pre-training and fine-tuning.

Since pre-training time is not proportional to the number of parameters\footnote{For example, in every training step only a very small portion of the embedding layer is updated.}, we measure the number of FLOPs (floating-point operations) required to do pre-training, as in \cite{clark2020electra}. FLOPs measure the number of mathematical operations performed on the hardware during the whole pre-training phase, thus this number is independent of the used accelerators (GPUs or TPUs) and of the model size.

\subsection{Pre-training}
\label{sec:pre_training}
We consider the same pre-training setting as BERT \cite{devlin2019bert} for the \textit{base} architectures. More precisely, we pre-train each model on the English Wikipedia and the BookCorpus dataset for $900K$ steps with a maximum sequence length of $128$ tokens and the last $100K$ steps at $512$. This setting allows for saving a lot of training time because the computational complexity of the attention layer is quadratic with the maximum sequence length. However, performance is slightly affected, but since we use the same pre-training setting (we also pre-train the baselines) for every model, this is a fair comparison.

We train every model with a learning rate equal to $1*10^{-4}$ and the Adam optimizer with the following parameters: $\epsilon = 1*10^{-8}, \beta_1 = 0.9, \beta_2 = 0.999$. The learning rate scheduler is designed to warm-up for $10K$ steps and then to decreases linearly. We used a batch size of $256$, and we apply a weight decay rate of $0.01$ over the network parameters to stabilize training.

Since ELECTRA models require more FLOPs because of the generator, we reduce the number of steps proportionally to the presence of the additional generator as in \cite{clark2020electra}. Then, we train for a total of $766K$ steps, of which $689K$ with a maximum sequence length of 128 tokens and the remaining $77K$ at 512.

\subsection{Fine-tuning}
This section provides the details about all the datasets and the experimental settings for fine-tuning.

\paragraph{GLUE}
The General Language Understanding Evaluation (GLUE) \cite{wang2019glue} is a well-known benchmark suite to test models in 9 different NLU tasks. This collection includes: (i) two datasets to test performance in paraphrasing capabilities, one composed of questions (QQP) pairs and the other of the sentence pairs (MRPC); (ii) a dataset for question-answer entailment (QNLI) derived from the SQUAD dataset \cite{rajpurkar2016squad}; (iii) three datasets for textual entailment (RTE, MNLI and WNLI); (iv) a single dataset (STS-B) to test the model on textual similarity; (v) a dataset (SST-2) to evaluate performance on sentiment analysis and finally (vi) a dataset to check linguistic acceptability (CoLA).

Results on the development set are computed by training with a batch size of $32$ for $3$ epochs, a learning rate of $2*10^{-5}$, a max sequence length of $128$ and by taking the \textit{average} over the $5$ runs with different seeds. For every model, we take the best checkpoint on the development set and we evaluate it on the GLUE Leader-board.

\paragraph{SQUAD}
The Stanford Question Answering Dataset \cite{rajpurkar2016squad} is a collection of questions created by crowd-workers and the relative article passage taken from Wikipedia in which there \textit{may} be the answer. The dataset features $100K$ training examples and slightly more than $10K$ validation examples. Since no test set is publicly available, we compare the models by training all of them for a fixed amount of steps with the same hyper-parameters. In particular, we train for $3$ epochs with a batch size of $32$ and a learning rate of $3*10^{-5}$. Moreover, we truncate input longer than $384$ tokens, and we repeat the experiment over each model for $3$ times with different seeds.

\paragraph{ASNQ-Reduced}
Answer-Sentence Natural Questions (ASNQ) \cite{garg2019tanda}, is a dataset built for the Answer Sentence Selection (AS2) tasks. It was built using the Natural Questions (NQ) corpus from \cite{kwiatkowski-etal-2019-natural}, which was originally created for Machine Reading. Natural Questions contains questions asked to the Google search engine and corresponding Wikipedia pages that almost always contains the answer. Crowd-workers extract long and short sentences from the articles. A long sentence is typically a paragraph or an HTML bounding box, while a short answer, which should be contained in the long, is a concise exact answer to the question.
The TandA \cite{garg2019tanda} identified four different type of labels as shown in Table \ref{tab:asnqc_dataset}

\begin{table}[h]
    \centering
    \small
    \begin{tabular}{c|c|c|c|c}
        Label & $S \in LA$ & $SA \in S$ & \# Train & \# Dev \\
        \midrule
        1 & No & No & 19,446,120 & 870,404 \\
        2 & No & Yes & 428,122 & 25,814 \\
        3 & Yes & No & 442,140 & 29,558 \\
        4 & Yes & Yes & 61,186 & 4,286 \\
    \end{tabular}
    \caption{\small $S$ is a sentence in the Wikipedia article, $SA$ is the short answer and $LA$ is the long answer.}
    \label{tab:asnqc_dataset}
\end{table}

The resulting dataset contains more than $20M$ entries, which is great for transfer learning. However, we remove all the sentences with the label $1$ because (i) they are easily recognizable since they are entirely off-topic; (ii) \cite{garg2019tanda} shows that transfer performance is not affected by this choice. By considering only sentences with labels equal to 2, 3 or 4, we built a dataset called ASNQ-Reduced, composed of slightly less than $1M$ training examples and about $60K$ validation samples. In order to have also a test set, we take the train, validation, and test splits from \cite{soldaini-moschitti-2020-cascade}, obtaining a validation and a test set with about $30K$ entries each.

We measure the performance using Mean Average Precision (MAP), Mean Reciprocal Rank (MRR) and Precision@1. For fine-tuning, we used a batch size of $128$, a learning rate of $2*10^{-5}$ and a maximum sequence length of $128$ tokens. We fine-tune $10$ times with different seeds.

\paragraph{WikiQA}
WikiQA \cite{yang2015wikiqa} is a dataset for Answer Sentence Selection built from questions asked to the Microsoft Bing search engine. Questions have been manually paired with answers taken from Wikipedia articles and labelled as relevant or not. The original WikiQA dataset is provided in different versions based on how questions with all negative or all positive answers are filtered. We follow the most common approach, excluding the questions with only negative answers for training and both questions with all negative or all positive answers for the development and test sets.
By applying those filters, we obtain $2,118$ questions, and $20,360$ answers for training, $121$ questions and $1,126$ answers for the development set and $237$ questions and $2,341$ answers in the test set. 
We train using batches with $32$ examples, a learning rate of $1*10^{-6}$ and for $10$ epochs. As with ASNQ-R, we measure performance with Mean Average Precision (MAP), Mean Reciprocal Rank (MRR) and Precision@1. For each metric, we take the average of over $10$ runs with different seeds to show reliable results.

\paragraph{ASNQ-R $\xrightarrow{}$ WikiQA}
We also test our models on transfer learning for the AS2 task. We adopt the same setting as TandA \cite{garg2019tanda} and we exploit the ASNQ-R dataset to do the transfer step, after which we fine-tune and test on WikiQA. Hyper-parameters for transfer and fine-tuning are the same used for fine-tuning ASNQ-R and WikiQA respectively.

\begin{table*}[h]
    \small
    \centering
    \begin{tabular}{l|cccccc}
        \toprule
        \textbf{Models} & RoBERTa-MLM & RoBERTa-RTS & RoBERTa-C-RTS & RoBERTa-SLM & ELECTRA & ELECTRA-SLM-all \\
        \textbf{FLOPS}  & $1.64*10^{19}$ & $1.54*10^{19}$ & $1.54*10^{19}$ & $1.64*10^{19}$ & $1.98*10^{19}$ & $2.55*10^{19}$ \\
        \bottomrule
    \end{tabular}
    \caption{\small FLOPS used to pre-train each model. The huge gap between ELECTRA and ELECTRA-SLM-all is due to the ELECTRA model using only a small binary classification head on the discriminator as opposed to ELECTRA enhanced with SLM, which does token prediction for every input. Moreover, even by reducing the number of training steps of the ELECTRA models by about the $25\%$ (to balance the presence of a discriminator with size $1/3$) it uses more FLOPs than RoBERTa models. Finally, notice that RoBERTa-RTS and -C-RTS use less memory thanks to having only a binary classification head.}
    \label{tab:flops}
\end{table*}

\begin{table*}[h]
    \small
    \centering
    \begin{tabular}{lccccccccccc}
        \toprule

        \multirow{2}{*}{\textbf{Model}} & \textbf{CoLA} & \textbf{MNLI} & \textbf{MRPC} & \textbf{QNLI} & \textbf{QQP} & \textbf{RTE} & \textbf{SST-2} & \textbf{STS-B} & \textbf{AVG} \\
        & matt & acc & acc & acc & acc & acc & acc & spear & \% & \\ 
        \toprule

        BERT-MLM+NSP    & 54.0 & 83.3 & 82.9 & 90.6	& 89.8 & 63.6 & 91.3 & 87.8 & 80.4 \\
        
        \midrule

        RoBERTa-MLM     & 55.1 & 83.1 & 84.7 & 89.7 & 90.2 & 57.8 & 91.1 & 87.0 & 79.8 \\
        RoBERTa-RTS     & 55.7 & 82.0 & 85.2 & 89.1 & 89.7 & 62.6 & 90.1 & 85.5 & 80.0 \\
        RoBERTa-C-RTS   & 57.3 & 81.4 & 83.7 & 89.2 & 89.5 & 62.2 & 89.7 & 85.8 & 79.9 \\
        RoBERTa-SLM     & 56.0 & 82.5 & 85.8 & 89.0 & 89.8 & 65.0 & 91.7 & 87.5 & \textbf{80.9} \\

        \midrule

        ELECTRA         & 60.9 & 83.3 & 86.0 & 90.7 & 90.8 & 69.5 & 90.8 & 88.2 & \textbf{82.5} \\
        ELECTRA-SLM-all & 51.4 & 82.5 & 84.2 & 89.3 & 90.4 & 61.4 & 90.8 & 87.1 & 79.6 \\

        \bottomrule
    \end{tabular}
    \caption{\small Results on GLUE development set as the average over $5$ different runs. We measure performance on STS-B and CoLA with Matthews and Spearman correlation coefficient respectively and on the other GLUE tasks with accuracy. The results average show a standard deviation that reaches up to $0.9\%$ because of the high standard deviation of results on small datasets like CoLA, RTE, SST-2 and STS-B. As in \cite{clark2020electra, devlin2019bert}, we do not show results on WNLI because it is even hard to beat a trivial majority classifier. We include also the BERT-\textit{base} model trained by \cite{devlin2019bert} using the same compute as our RoBERTa-MLM to show that our models are aligned with the state-of-the-art. The NSP loss, as reported in many works, does not provide a noticeable improvement \cite{liu2019roberta}.}
    \label{tab:glue_dev}
\end{table*}

\begin{table*}[h]
    \small
    \centering
    \begin{tabular}{lcccccccccccc}
        \toprule

        \multirow{2}{*}{\textbf{Model}} & \textbf{CoLA} & \textbf{MNLI} & \textbf{MRPC} & \textbf{QNLI} & \textbf{QQP} & \textbf{RTE} & \textbf{SST-2} & \textbf{STS-B} & \textbf{WNLI} & \textbf{AVG} \\
        & matt & acc & acc & acc & acc & acc & acc & spear & acc & \% \\ 
        \toprule

        BERT-MLM+NSP    & 50.0 & 82.9 & 82.7 & 90.4 & 88.3 & 65.2 & 90.6 & 82.8 & 65.8 & \textbf{77.6} \\

        \midrule

        RoBERTa-MLM     & 54.7 & 82.6 & 84.4 & 90.2 & 88.7 & 55.6 & 91.2 & 82.5 & 60.3 & 76.7 \\
        RoBERTa-RTS     & 56.2 & 81.6 & 84.2 & 89.6 & 87.7 & 61.0 & 90.3 & 79.5 & 65.1 & 77.2 \\
        RoBERTa-C-RTS   & 52.7 & 81.6 & 84.0 & 89.5 & 88.7 & 61.4 & 89.5 & 79.6 & 65.8 & 77.0 \\
        RoBERTa-SLM     & 54.8 & 82.7 & 81.9 & 89.5 & 88.7 & 61.3 & 91.0 & 83.6 & 65.1 & \textbf{77.6} \\

        \midrule

        ELECTRA         & 59.6 & 82.9 & 85.3 & 91.0 & 89.3 & 66.1 & 91.7 & 84.8 & 65.1 & \textbf{79.5} \\
        ELECTRA-SLM-all & 51.1 & 82.6 & 84.9 & 89.5 & 89.1 & 62.9 & 92.6 & 82.3 & 65.1 & 77.8 \\

        \bottomrule
    \end{tabular}
    \caption{\small Results on GLUE test set. We evaluate with the same standard metrics used on the development set. For each task we fine-tune $5$ times and take the best model on the development set. Again, we include results on BERT-\textit{base} of \cite{devlin2019bert} as described in Table \ref{tab:glue_dev}.}
    \label{tab:glue_test}
\end{table*}

\section{Results}
\label{sec:results}
As said in the previous section, we evaluated our models on GLUE, SQUAD, WikiQA and ASNQ-R. We also include some baselines of RoBERTa-base and ELECTRA-base. To fairly compare the various objectives, we trained all baselines using the setting described in Section \ref{sec:pre_training}. For this reason, these baselines cannot be compared directly with the models released by the RoBERTa or ELECTRA teams. Moreover, we do not use tricks to improve results on GLUE as many recent works do \cite{lan2020albert, clark2020electra, liu2019roberta, yang2020xlnet}. Finally, it is worth mentioning that we reproduced the performance of ELECTRA in a standard pre-training setting, without using their custom (i) optimizer\footnote{See \url{https://github.com/google-research/electra/blob/8a46635f32083ada044d7e9ad09604742600ee7b/model/optimization.py\#L70}.}, (ii) very high and unstable learning rate, and (iii) the particular layer-wise learning rate decay \cite{clark2020electra}. This way, we have highly improved reproducibility. Baselines are reported as \textbf{RoBERTa-MLM} and \textbf{ELECTRA} in the results tables.

Before discussing results, we show the number of FLOPs used to pre-train each model in Table \ref{tab:flops}. We measure FLOPs using the same procedure explained in the \cite{clark2020electra}. These numbers are significant indicators but may reflect in different practical performance if the underlying hardware implements special acceleration for some type of matrix operations.

\begin{table*}[t]
    \small
    \centering
    \begin{tabular}{l|cc|ccc|ccc|ccc}
        \toprule

        \multirow{2}{*}{\textbf{Model}} & \multicolumn{2}{c|}{\textbf{SQUAD V1.1}} & \multicolumn{3}{c|}{\textbf{WikiQA}} & \multicolumn{3}{c|}{\textbf{ASNQ-R}} & \multicolumn{3}{c}{\textbf{ASNQ-R} $\xrightarrow{}$ \textbf{WikiQA}} \\
        & EM & F1 & MAP & MRR & P@1 & MAP & MRR & P@1 & MAP & MRR & P@1 \\

        \toprule

        RoBERTa-MLM     & 80.8 & 88.1 &   73.3 & 74.0 & 62.2 &   79.1 & 84.0 & 74.4 &   \textbf{83.4} & 84.1 & \textbf{74.4} \\
        RoBERTa-RTS     & 78.7 & 86.2 &   74.3 & 75.1 & 64.4 &   78.4 & 83.3 & 73.6 &   83.2 & 83.8 & 72.6 \\
        RoBERTa-C-RTS   & 79.2 & 86.7 &   74.3 & 75.2 & 64.8 &   78.6 & 83.4 & 73.9 &   83.3 & \textbf{84.2} & 73.9 \\
        RoBERTa-SLM     & \textbf{81.2} & \textbf{88.7} &   \textbf{75.5} & \textbf{76.0} & \textbf{65.1} &   \textbf{79.4} & \textbf{84.1} & \textbf{74.8} &   82.5 & 83.0 & 72.5 \\

        \midrule

        ELECTRA         & \textbf{81.3} & \textbf{88.7} &   \textbf{75.6} & \textbf{75.9} & 63.7 &   \textbf{80.2} & \textbf{84.9} & \textbf{76.0} &   \textbf{84.5} & \textbf{85.2} & \textbf{75.0} \\
        ELECTRA-SLM-all & 80.5 & 88.1 &   75.2 & 75.6 & \textbf{64.6} &   79.1 & 83.9 & 74.2 &   83.5 & 83.9 & 73.7 \\

        \bottomrule
    \end{tabular}
    \caption{\small Results on development sets for SQUAD v1.1, WikiQA, ASNQ-R and WikiQA after the transfer step on ASNQ-R. Standard deviations for SQUAD results are always below $0.3\%$. The standard deviations for ASNQ are instead always smaller than $0.5\%$ for MAP and MRR. WikiQA, being a smaller dataset, provides less uniform results and the standard deviation reaches $1\%$ for MAP and MRR on all models. In ASNQ-R $\xrightarrow{}$ WikiQA, the standard deviation is even higher because in transfer learning one combines two training phases. Thus, considering again MAP and MRR, the standard deviation increases up to $1.7\%$ for RoBERTa-RTS and RoBERTa-SLM, while it is slightly over $1.0\%$ for the other models.}

    \label{tab:qa_results_dev}
\end{table*}

\begin{table*}[t]
    \small
    \centering
    \begin{tabular}{l|ccc|ccc|ccc}
        \toprule

        \multirow{2}{*}{\textbf{Model}} & \multicolumn{3}{c|}{\textbf{WikiQA}} & \multicolumn{3}{c|}{\textbf{ASNQ-R}} & \multicolumn{3}{c}{\textbf{ASNQ-R} $\xrightarrow{}$ \textbf{WikiQA}} \\
        & MAP & MRR & P@1 & MAP & MRR & P@1 & MAP & MRR & P@1 \\

        \toprule

        RoBERTa-MLM     & 70.8 & 72.1 & 58.6 &   78.7 & \textbf{83.3} & 73.0 &   82.9 & 84.2 & 74.3 \\
        RoBERTa-RTS     & 71.0 & 72.0 & 57.7 &   78.2 & 82.9 & 72.5 &   83.0 & 84.5 & 74.6 \\
        RoBERTa-C-RTS   & 71.0 & 72.1 & 57.6 &   78.4 & 83.1 & 72.8 &   \textbf{84.3} & \textbf{85.8} & \textbf{77.1} \\
        RoBERTa-SLM     & \textbf{72.1} & \textbf{73.2} & \textbf{59.6} &   \textbf{78.8} & \textbf{83.3} & \textbf{73.3} &   84.2 & 85.4 & 75.4 \\

        \midrule

        ELECTRA         & \textbf{75.3} & \textbf{76.7} & \textbf{62.8} &   \textbf{79.3} & \textbf{83.8} & \textbf{73.9} &   86.9 & 88.0 & 80.4 \\
        ELECTRA-SLM-all & 72.1 & 73.4 & 59.8 &   79.0 & 83.5 & 73.7 &   \textbf{87.4} & \textbf{88.9} & \textbf{82.5} \\

        \bottomrule
    \end{tabular}
    \caption{\small Results on the test set for WikiQA, ASNQ-R and WikiQA after the transfer step on ASNQ-R. Standard deviations of MAP and MRR for WikiQA are always below $0.8\%$ apart from RoBERTa-C-RTS and ELECTRA that reach $1.4\%$. In the case of ASNQ-R, standard deviations are smaller for most models $<0.3\%$ apart from ELECTRA, which reaches $1.5\%$ for both MAP and MRR. Finally, standard deviations for ASNQ $\xrightarrow{}$ WikiQA are small in average and are always below $0.5\%$. This proves that transfer learning is also a way to stabilize results on the target task.}
    \label{tab:qa_results_test}
\end{table*}

Tables \ref{tab:glue_dev} and \ref{tab:glue_test} show the main results on the development and test sets of GLUE. They show that all the considered approaches, with the exception of ELECTRA base, obtain comparable performance. In particular, the RoBERTa models based on RTS and C-RTS obtain a general improvement over MLM on the test set while requiring a lower amount of computational effort to be pre-trained. Besides, is also noticeable that the difference in performance between ELECTRA and our techniques is not significant on the most stable tasks of GLUE (QQP and MNLI). Indeed, ELECTRA widely outperforms our models only on CoLA, RTE and STS-B.

It is also worth mentioning that our RoBERTa-SLM model achieves better performance than MLM on both GLUE development and test sets by simply removing the MASK token from the pre-training. On the other hand ELECTRA-SLM-all struggles in every GLUE test, where the data availability is scarce, thus showing low generalization capabilities. Moreover, it requires more computation resources due to the large classification head on the main network\footnote{The equivalent of the discriminator in ELECTRA, but is not a discriminator.}. This confirms that the ELECTRA discriminator works well even if it is not trained with a generative approach.

Table \ref{tab:qa_results_dev} and Table \ref{tab:qa_results_test} show the results on the different QA and AS2 benchmarks described in Section \ref{sec:experiments}. The results on these datasets are aligned with those on the GLUE benchmark.

In particular, considering the RoBERTa models, SLM is always superior to MLM for every considered dataset and it even matches ELECTRA on SQUAD. It obtains a significant margin over MLM especially in low-resource tasks like WikiQA, confirming better generalization capabilities. Simultaneously, RTS and C-RTS objectives lead to similar performances but require a lower amount of resources to be pre-trained. Moreover, RoBERTa-C-RTS shows that a harder pre-training using challenging replacements provides better performances in transfer learning, where RoBERTa-RTS is very similar to RoBERTa-MLM instead.

Besides studying the results of the ELECTRA models, we noticed wider differences between the two approaches. In particular, considering the results on the test sets, the SLM-all are comparable with the ELECTRA performance on the ASNQ-R benchmark ($79.0$ of ELECTRA-SLM-all against $79.3$ considering MAP), worse on the WikiQA ($75.3$ and $72.1$) but better after the transfer learning step from ASNQ-R to WikiQA ($87.4$ of the SLM-all approach compared with the $86.9$ of ELECTRA). These results point out the quality of SLM-all applied to an ELECTRA model, but also its limits. Indeed, ELECTRA-SLM-all demonstrates that the advantages of transforming the ELECTRA discriminator into a generative network is expensive and does not ensure better performance for the majority of the tasks, but could be still a valuable choice for transfer learning.

\section{Conclusion and Future Works}
\label{sec:conclusion}

In this work, we studied several methods to efficiently pre-train Transformer models. Our approaches aim to match the results of well known pre-training objectives such as MLM but consuming less computational resources. These research directions have several benefits, since a lower computational effort leads to a shorter training and to lower memory usage. We evaluate our approach using several benchmark dataset such as GLUE, SQUAD, WikiQA and ASNQ-R to have a clear understanding of the behaviours of our models.

The results show that RoBERTa-RTS and RoBERTa-C-RTS can match the performance of the vanilla RoBERTa-MLM in every task while requiring less training time. Moreover, RoBERTa-C-RTS shows better performance also with respect to RTS and MLM in transfer learning. RoBERTa-SLM is another valid alternative to MLM: while using exactly the same compute, it outperforms MLM on every task by a consistent margin, thus confirming that the removal of the MASK token is essential. Furthermore, RoBERTa-SLM is able to match ELECTRA in some tasks and confirms that the original ELECTRA benefits also from a fine-grained hyper-parameters search, other than using the whole output to compute the loss and no MASK token. Finally, our ELECTRA-SLM-all model shows that the discriminator of ELECTRA is not affected by doing only binary classification, but has interesting performance in transfer-learning, reaching the highest performance in this work.

We plan to combine our new techniques with efficient architectures like DistilBERT and ALBERT in the near future, to take advantage of both a lighter structure and a more effective pre-training objective.

\bibliographystyle{unsrt}  
\bibliography{references}  

\end{document}